\pdfoutput=1
\documentclass[letterpaper, 10 pt, journal, twoside]{./support/IEEEtran}

\usepackage{times}
\usepackage[pdftex]{graphicx}
\usepackage{subfigure}
\usepackage{amsmath,amssymb,amsopn,amstext,amsfonts}
\usepackage{cancel}
\usepackage[space]{cite}
\usepackage{soul}
\usepackage{cuted}
\usepackage{lipsum,multicol}
\usepackage{stfloats}
\everymath{\displaystyle}
\usepackage{booktabs}
\usepackage{threeparttable}
\usepackage{amsthm}
\usepackage{xfrac}
\usepackage{balance}
\usepackage{color}
\usepackage{mathtools}

\usepackage{tabularx}
\usepackage{bm}
\usepackage{diagbox}
\usepackage{float}
\usepackage{epstopdf}
\usepackage{url}
\usepackage{multirow}
\usepackage{multicol}
\usepackage{tikz}
\usepackage[linkcolor=black,citecolor=black,urlcolor=black,colorlinks=true]{hyperref}
\usepackage{enumitem}
\usepackage[ruled, vlined, linesnumbered, nofillcomment]{algorithm2e}
\usepackage{pifont}
\usepackage{siunitx}

\DeclareMathAlphabet\mathbfcal{OMS}{cmsy}{b}{n}

\usepackage[normalem]{ulem}

\let\oldFootnote\footnote
\newcommand\nextToken\relax

\renewcommand\footnote[1]{%
    \oldFootnote{#1}\futurelet\nextToken\isFootnote}
\newcommand\isFootnote{%
    \ifx\footnote\nextToken\textsuperscript{,}\fi}

\definecolor{celadon}{rgb}{0.78, 0.93, 0.80}
\pagecolor{celadon}
\nopagecolor
\soulregister\cite7
\soulregister\citep7
\soulregister\citet7
\soulregister\ref7
\soulregister\it7
\soulregister\pageref7

\usepackage{arydshln}
\graphicspath{{figures/}}
\DeclareGraphicsExtensions{.pdf,.png,.jpg,.eps}

\IEEEoverridecommandlockouts

\title{D-BDM: A Direct and Efficient Boundary-Based Occupancy Grid Mapping Framework for LiDARs}

\author{
    Benxu Tang,
    Yixi Cai,
    Fanze Kong,
    Longji Yin,
    Fu Zhang
\thanks{Benxu Tang, Fanze Kong, Longji Yin and Fu Zhang are with Mechatronics and Robotic Systems (MaRS) Laboratory, Department of Mechanical Engineering, The University of Hong Kong (email: tangbenx@connect.hku.hk; kongfz@connect.hku.hk; ljyin@connect.hku.hk; fuzhang@hku.hk). Yixi Cai is with Department of Robotics, Perception, and Learning, KTH Royal Institute of Technology (email: yixica@kth.se)}
}

\begin{document}
\maketitle
\begin{abstract}

	Efficient and scalable 3D occupancy mapping is essential for autonomous robot applications in unknown environments. However, traditional occupancy grid representations suffer from two fundamental limitations. First, explicitly storing all voxels in three-dimensional space leads to prohibitive memory consumption. Second, exhaustive ray casting incurs high update latency. 
    A recent representation alleviate memory demands by maintaining only the voxels on the two-dimensional boundary, yet they still rely on full ray casting updates.
    This work advances the boundary-based framework with a highly efficient update scheme. We introduce a truncated ray casting strategy that restricts voxel traversal to the exterior of the boundary, which dramatically reduces the number of updated voxels. In addition, we propose a direct boundary update mechanism that removes the need for an auxiliary local 3D occupancy grid, further reducing memory usage and simplifying the map update pipeline. We name our framework as D-BDM.
    Extensive evaluations on public datasets demonstrate that our approach achieves significantly lower update time and reduced memory consumption compared with the baseline methods, as well as the prior boundary-based approach.
	
\end{abstract}


\section{Introduction}
\begin{figure}[t] 
    \centering
    \includegraphics[width=\linewidth]{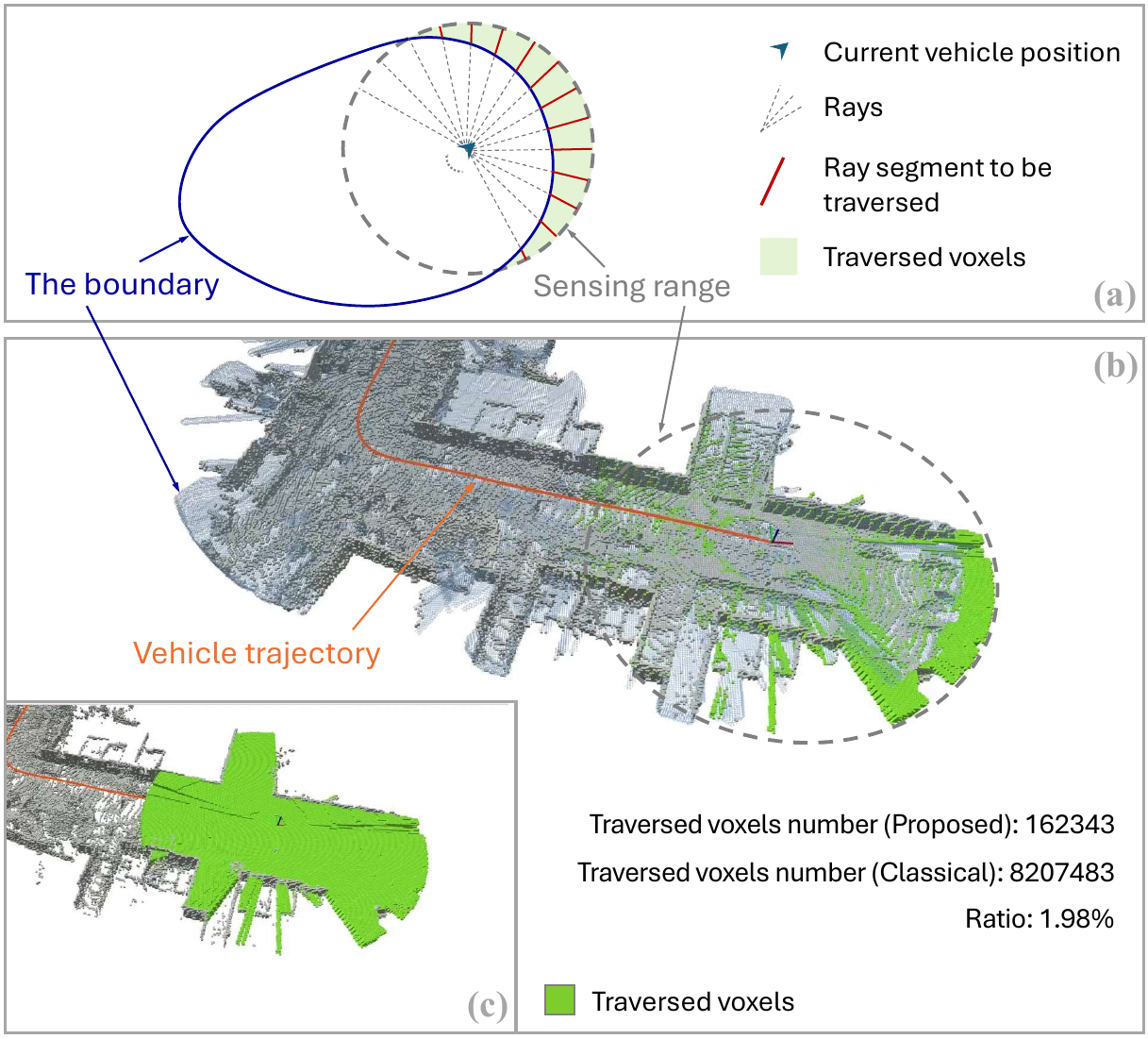}
    \vspace{-10pt} 

    \caption{(a) Illustration of the proposed update scheme. Ray traversal is limited to segments outside the boundary, avoiding full-ray processing. (b) Visualization of the experiment in \textit{kitti\_00} sequence, showing the boundary voxels (transparent blue) and the traversed voxels (green) of the proposed method. Occupied voxels are highlighted in grey. (c) Comparison with classical full ray casting, demonstrating that the proposed method greatly reduces the number of traversed voxels (i.e., down to 1.98\%).}

    \label{fig:cover}
    \vspace{-10pt}
\end{figure}

Autonomous robots are increasingly deployed in search-and-rescue~\cite{rouvcek2020darpa, tranzatto2022cerberus}, emergency response~\cite{kawatsuma2012emergency, seungsub2017study}, autonomous inspection and reconstruction~\cite{yoder2016autonomous, tabib2021autonomous}, where they must operate reliably in complex and previously unknown environments. 
Numerous planning and perception algorithms~\cite{kong2021avoiding, ren2025safety, yang2022far, dang2019graph, zhou2021fuel, tang2023bubble} have been proposed in recent years to enable safe navigation and efficient task execution.
Fundamental to these capabilities is the determination of the occupancy state of the environment—classifying regions as free, occupied, or unknown—so that robots can make informed and safety-ensured decisions in real time.

Occupancy grid maps offers an effective solution by discretizing the environment into a voxelized grid. Each voxel encodes an independent occupancy status. Determining the occupancy state of any location in the environment can be achieved by querying the corresponding voxel in the grid. This representation provides a simple and efficient representation of the environment, making it widely adopted in robotic systems.

However, occupancy grid mapping faces two fundamental challenges: memory consumption and update efficiency. Traditional occupancy grid maps maintain all voxels in the three-dimensional (3D) grid. Some maintain these voxels using contiguous array or hash-table~\cite{moravec1996robot, ren2023rog, niessner2013real}, others organize these voxels by hierarchical octrees~\cite{hornung2013octomap, duberg2020ufomap, cai2023occupancy}. However, as the environment grows in scale or map resolution increases, the number of grid voxels that need to maintain increase significantly, these methods leading to prohibitive memory consumption. To alleviate this issue, a recent work~\cite{BDM} propose to store only the two-dimensional (2D) voxels on the boundary, referred to as BoundaryMap. This low-dimensional representation greatly reduces memory usage.
{However, BoundaryMap still relies on a dense local 3D grid map during map updates, which can contribute a non-negligible fraction of the overall memory usage.}

In terms of map updates, the ray casting technique is widely adopted in traditional occupancy grid mapping methods as well as the BoundaryMap. Specifically, each point in sensor scan updates the corresponding voxel towards occupied, and a ray is casted from each sensor point to sensor origin, where all voxels along the ray are updated toward free state. 
Ray casting process is computational expensive. A vast number of voxels are traversed and updated in the process, especially with high-density and long-range sensors (e.g., LiDARs). This leads to high update latency and preventing real-time performance.

To address these challenges, we leverage the boundary map framework and propose a novel method that significantly improve update efficiency while further reducing memory consumption.
Our main contributions are summarized as follows:
\begin{itemize}
\item Leveraging the properties of the boundary map, we propose a truncated ray casting strategy that greatly reduces the number of traversed voxels during map updates, thereby substantially improves map update efficiency.

\item We develop an efficient direct boundary map update scheme that eliminates the need for a local occupancy grid map, which further reduces memory consumption and simplifying the map update pipeline.

\item We conducted comprehensive benchmark experiments against state-of-the-art baselines to validate the proposed methods in terms of update efficiency and memory consumption.

\end{itemize}

\section{Related Works}
In this section, we review previous studies on occupancy grid mapping methods, focusing on their representation and map update methods.
\subsection{Occupancy Grid Maps}
{Occupancy grid maps aims to maintain the occupancy state of the environment, typically classifying locations as free, occupied, or unknown. To support efficient query and update operations, this information must be organized using appropriate map representations.}
Classical occupancy grid map representations can be broadly categorized into grid-based and octree-based methods, depending on the underlying data structure. For \textbf{grid-based} methods, \cite{roth1989building, elfes1995robot, moravec1996robot} represent the entire environment by a uniform grid. All voxels in the uniform grid are mapped into a contiguous memory block (i.e., an array). This design supports constant-time ($\mathcal{O}(1)$) for a single voxel operation (i.e., query or modify the occupancy status of a voxel) but incurs prohibitive memory usage at large scales. \cite{niessner2013real} improves memory efficiency by storing only observed voxels (free or occupied) in a hash table, rather than maintaining the full grid. However, hash collisions can degrade worst-case operation time to $\mathcal{O}(n)$. Moreover, at fine resolutions or over large environments, even storing all free and occupied voxels still remains memory-intensive. To mitigate these issues, \textbf{octree-based}~\cite{hornung2013octomap, duberg2020ufomap} approaches recursively subdivide the environment, merging regions with homogeneous occupancy status to reduce memory demand. However, the trade-off is that each voxel operation requires traversing the tree from root to leaf, resulting in logarithmic complexity, $\mathcal{O}(\log n)$, where $n$ is the tree depth.

Both grid- and octree-based occupancy maps explicitly maintain three-dimensional (3D) volumetric occupancy information, their memory demands still remain prohibitive as map resolution or environment scale further increase. To alleviate this issue, a recent work~\cite{BDM} proposes storing only the boundary of the observed volume as a two-dimensional (2D) closed surface, namely BoundaryMap. This dimensionality reduction substantially lowers memory usage. Then, the method provides a mechanism to determine the occupancy of any locations in 3D space directly from the maintained 2D boundary
{However, BoundaryMap introduces a grid-based 3D local map in its map update framework. To properly integrate all sensor measurements, the local map must fully cover the sensor’s field of view. As a result, the local map can account for a non-negligible portion of the overall memory consumption.}

\subsection{Update Methods}
Ray casting is a common technique for updating occupancy grid maps. In particular, for each measurement point in a sensor scan, a ray is traced from the sensor origin to the point. All voxels traversed by the ray are updated toward the free state, and the voxel at the endpoint is updated toward occupied. Despite its simplicity, this procedure traverses a substantial number of voxels and is computationally expensive. In grid-based maps, the complexity of ray casting is $\mathcal{O}\left(p\frac{R}{d}\right)$~\cite{cai2023occupancy}, where $p$ is the number of scan points, $R$ is the sensing range, and $d$ is the map resolution. {In BoundaryMap, map updates also involves a full ray casting over the grid-based local map, resulting in similar computational complexity.}
Octree-based maps incur an additional logarithmic factor due to tree traversal, yielding $\mathcal{O}\left(p\frac{R}{d}\log n\right)$ ~\cite{cai2023occupancy}. Several methods have attempted to reduce this cost. For example,~\cite{wurm2011hierarchies} employs a multi-resolution strategy that maintains submaps at coarser resolutions. \cite{duberg2020ufomap} downsamples scan points based on map resolution. However, these approaches modify the standard ray casting procedure and introduce discrepancies that degrade accuracy.

A recent work, namely D-Map~\cite{cai2023occupancy}, avoids the ray casting process that commonly used in classical methods. Instead, it updates occupancy using depth images, allowing occupancy states to be determined at higher-level octree cells without traversing all finest-resolution voxels. Furthermore, grid cells with determined states are removed at each update, yielding a decremental update behavior that reduces revisits. These designs substantially improve update efficiency. Nevertheless, because D-Map still employs an octree as its underlying representation, it continues to suffer from the inherent logarithmic overhead associated with the octree-based structure. In addition, the method also introduces several approximations that lead to noticeable accuracy degradation. Moreover, D-Map also cannot handle dynamic environments; once a voxel is marked occupied, subsequent observations cannot revert it to free. This limitation restricts its applicability in real-world scenarios.

To address these limitations, we advance the BoundaryMap framework by introducing an efficient update method, while also eliminating the need for the grid-based local map.
As a result, our approach achieves markedly superior update performance and lower memory usage compared with existing methods, including classical grid- and octree-based methods, D-Map, and BoundaryMap.

\section{Methdology}
\label{sec:methdology}

\begin{figure*}[t] 
    \centering
    \includegraphics[width=\textwidth]{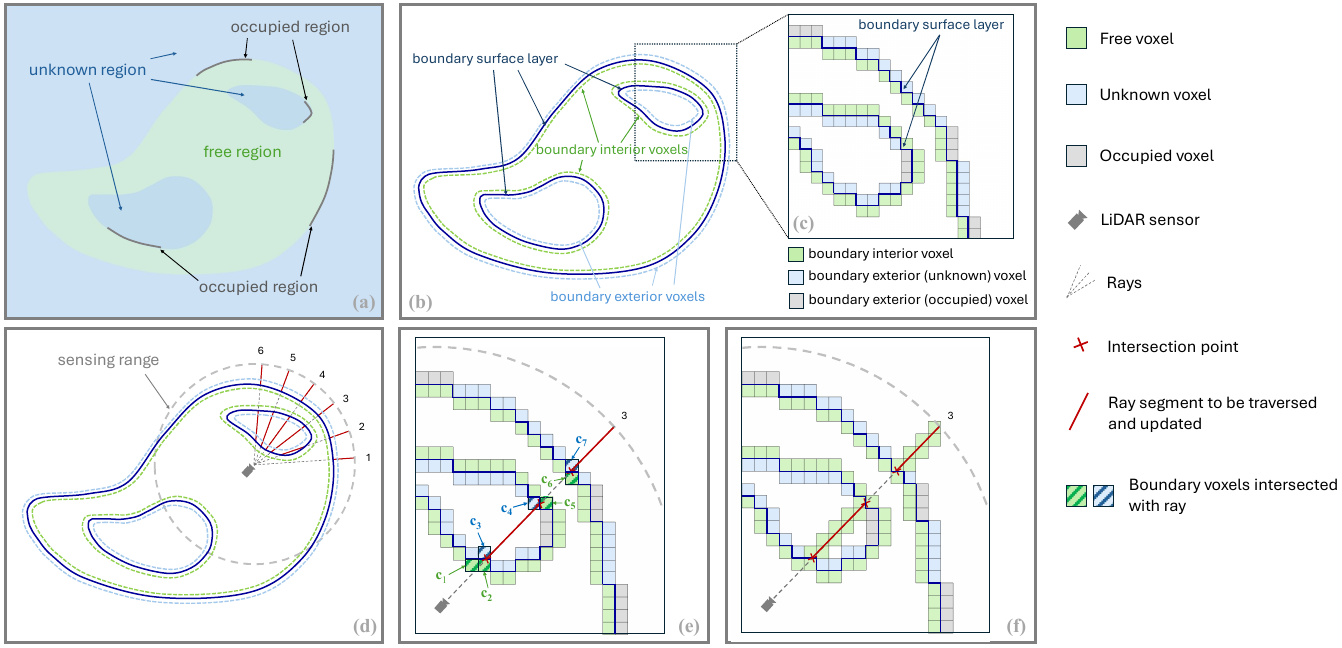}
    \vspace{-10pt} 

    \caption{(a) An uniform occupancy grid with free, unknown, and occupied regions, represented by green, blue, and grey, respectively.
    (b) The boundary surface layer is highlighted in dark blue, which separates the free regions from the adjacent occupied or unknown regions. The boundary map only represents voxels neighboring the boundary surface layer, which includes boundary interior voxels and boundary exterior voxels. (c) Detailed voxel-level view of (b). The boundary interior voxel having an occupancy state of free. The boundary exterior voxel is further classified as boundary exterior (unknown) voxel and boundary exterior (occupied) voxel, whose occupancy states are unknown and occupied, respectively. (d) Truncated ray casting process, where the ray traversal is only preformed at ray segments that located at the exterior of boundary surface layer. Such segments are highlighted in red. (e) A detailed voxel-level view of the truncated ray casting process of ray no.3. The intersection point of the ray and boundary surface layer is marked by red cross symbol. The boundary voxels that intersected with the ray are also highlighted. (f) Voxels traversed by these exterior ray segments are updated to free state.}

    \label{fig:boundary_map}
\end{figure*}

In this section, we first present our representation of occupancy information together with our rules in determining the occupancy states (see Section~\ref{sec:represent}). Then, {we introduce the truncated ray casting strategy in Section~\ref{sec:ray_cast} and the direct boundary map update scheme in Section~\ref{sec:upd_boundary}.}
\subsection{Occupancy State Representation and Determination}
\label{sec:represent}
Inspired by the BoundaryMap~\cite{BDM}, we represent the 3D free space using only its 2D boundary. As illustrated in {Fig.~\ref{fig:boundary_map}}, the boundary of the free space forms a closed surface, defined as the boundary surface layer.  To encode this surface, we maintain voxels that ``straddle" it, which are referred to as the boundary voxels. The boundary voxels consist of three categories: The first type is the voxels labeled as free that have at least one voxel within its 6-neighbors having non-free occupancy state (i.e., unknown or occupied). These voxels lie at the interior are classified as boundary interior voxels ($\mathbf{b}_{\mathrm{int}}$). Second, voxels labeled as unknown that has at least one free voxel within its 6-neighbors are classified as boundary exterior (unknown) voxels ($\mathbf{b}_{\text{ukn}}$). Finally, all voxels labeled as occupied are classified as boundary exterior (occupied) voxels ($\mathbf{b}_{\text{occ}}$). The latter two categories are collectively referred to as boundary exterior voxels ($\mathbf{b}_{\text{ext}}$). An illustration is presented in Fig.~\ref{fig:boundary_map}(c).

Traditional occupancy grid mapping methods adopt probabilistic updates to integrate new LiDAR measurements. Specifically, each voxel in the grid stores an occupancy probability value. At each scan, a ray is cast from the sensor origin to the LiDAR point in scan. If the ray pass through a voxel, its occupancy probability is decreased. Conversely, If a voxel contains LiDAR point(s), its occupancy probability is increased. The occupancy state of each voxel is then determined as free or occupied by thresholding its occupancy probability. Such probabilistic update scheme supports sensor noise and dynamic environments handling.
Given the high accuracy of LiDAR measurements, we discard this probabilistic rule and instead represent each voxel using a discrete occupancy state: free, occupied, or unknown. When a {LiDAR} point falls within a voxel, the voxel is directly marked as occupied. Conversely, if a ray passes through a voxel, the voxel is marked as free. However, due to the inaccuracy from grid discretization, a voxel may simultaneously be traversed by rays and hit by one or more LiDAR points within a single sensor scan. In such cases of conflict, we assign the voxel as occupied. 
Notably, this update policy also supports dynamic environments handling. A voxel previously labeled as occupied or free can be updated into the other state by subsequent measurements.

\subsection{Truncated Ray Casting}
\label{sec:ray_cast}

\begin{algorithm}[t]
\caption{Map Update Framework}
\small
\label{alg:boundary_update}
\DontPrintSemicolon
\KwIn{Sensor origin $\mathbf{x}$, sensor FoV $\mathcal{V}$, sensor scan $\mathcal{C}$, depth image resolution $\psi$}

\BlankLine
$\mathcal{V}^\text{2D} \gets \mathtt{GetProjectionArea}(\mathcal{V})$\;
$\mathcal{K} \gets \mathtt{UpdateCellsList}(\mathcal{V}^\text{2D} )$\;
$\mathbf{P} \gets \mathtt{GetBoundaryVoxelsInFoV}(\mathcal{K}, \mathcal{V})$\;
$\mathbf{L} \gets \mathtt{TruncatedRayCasting}(\mathbf{P}, \mathbf{x}, \mathcal{C}, \psi)$\;
$\mathtt{UpdateBoundaryMap}(\mathbf{L}, \mathcal{K})$\;

\BlankLine

\SetKwFunction{FTruncate}{TruncatedRayCasting}
\SetKwProg{Fn}{Function}{:}{}
\Fn{\FTruncate{$\mathbf{P}, \mathbf{x}, \mathcal{C}, \psi$}}{
$\mathcal{M} \gets \mathtt{GenerateDepthImage}(\mathcal{C}, \psi)$\;
$\mathtt{ProjectBoundaryVoxels}(\mathbf{P}, \mathcal{M}, \mathbf{x})$\;

\For{pixel $\mathtt{p}_i \in \mathcal{M}$}{
    $\mathbf{T}_i \gets \mathtt{RetrieveData}(\mathtt{p}_i)$\;
    $\mathtt{RemoveNotIntersect}(\mathbf{T}_i)$\;
    $\mathtt{SortByDepth}(\mathbf{T}_i)$\;
    $\mathtt{s} \gets \textit{interior} $\;

    \For{$\mathbf{t}_k \in \mathbf{T}_i$}{
        \If{$\mathtt{s} = \textit{interior}$}{
        \If{$\mathbf{t}_k = \mathbf{b}_{\mathrm{ext}}$}{
        $\mathtt{s} \gets \textit{exterior}$\;
        $\mathtt{RayCastStart}()$\;
        $\mathbf{L} \gets \mathtt{RecordUpdatedVoxels}()$\;}
        }
        \ElseIf{$\mathtt{s} = \textit{exterior}$}{
        \If{$\mathbf{t}_k = \mathbf{b}_{\mathrm{int}}$}{
        $\mathtt{s} \gets \textit{interior}$\;
        $\mathtt{RayCastEnd}()$\;}
        }
    }
}
\Return{$\mathbf{L}$}
}

\BlankLine

\SetKwFunction{FUpdate}{UpdateBoundaryMap}
\SetKwProg{Fn}{Function}{:}{}
\Fn{\FUpdate{$\mathbf{L}, \mathcal{K}$}}{

$\mathcal{F} \gets \varnothing$\;
\For{voxel $\mathbf{n} \in \mathbf{L}$}{
    $\mathcal{F} \gets \mathcal{F} \cup \{\mathbf{n}\}$\;
    $\mathcal{F} \gets \mathcal{F} \cup \mathtt{SixNeighbors}(\mathbf{n})$\;
}

\For{2D grid cell $\mathbf{k} \in \mathcal{K}$}{
    $\mathbf{B}_{{k}} \gets \mathtt{GetBoundaryVoxels}(\mathbf{k})$\;
    \For{voxel $\mathbf{b}_{{k}} \in \mathbf{B}_{{k}}$}{
        \If{$\mathbf{b}_{{k}} \in \mathcal{F}$}{
            $\mathtt{Remove}(\mathbf{b}_{{k}})$\;
        }
    }
}

\For{voxel $\mathbf{m} \in \mathcal{F}$}{
    $\mathtt{ComputeBoundaryStatus}(\mathbf{m})$\;
    \If{$\mathtt{isBoundary}(\mathbf{m})$}{
        $\mathtt{Add}(\mathbf{m})$\;
    }
}
}
\end{algorithm}

{To enable efficient map updates, we propose a truncated ray casting process that operates only at the exterior of the boundary. Specifically, only voxels in this exterior region are traversed and updated to free states. Those within the boundary are already classified as free, thus traversing them is unnecessary. Specifically,} a ray is emitted from the sensor origin toward each LiDAR return in the scan. The ray may intersect the boundary surface layer at one or multiple points, referred to as intersection points. These intersections points are then used to determine the segments of the ray that lies outside the boundary surface layer. Only voxels traversed by these exterior ray segments are updated toward the free state, instead of traversing the entire ray. {This design substantially reduces the number of voxels traversed during ray casting.} The complete procedure is detailed below and summarized in Alg.~\ref{alg:boundary_update} (Function \texttt{TruncatedRayCasting}, Lines 6--24).

First, to efficiently identify the intersection points, we adopt a depth-image rasterization strategy inspired by~\cite{cai2023occupancy}. A depth image is first constructed, with its resolution, denoted as \(\psi\), determined by the angular resolution of the LiDAR. The LiDAR scan is represented by this depth image, where each pixel stores the depth value of the corresponding LiDAR return. This process is referred to as \texttt{GenerateDepthImage} (Line 7). Subsequently, boundary voxels within the LiDAR field of view (FoV) are also projected onto the same image, namely \texttt{ProjectBoundaryVoxels} (Line 8). Specifically, for a boundary voxel \(\mathbf{c}\) with center coordinates \((x_c, y_c, z_c) \in \mathbb{R}^3\) in the Cartesian frame, we first convert its position to spherical coordinates \((r_c, \theta_c, \phi_c) \in \mathbb{R}^3\). The projection accounts for the voxel’s spatial extent by using its circumscribed circle: given the voxel edge length \(d\), the diameter of the circumscribed circle is \(\sqrt{2}d\). As a result, the projection area on the depth image is enclosed by a rectangular region, whose bounding vertices \(\mathbf{p}_{\min}\) and \(\mathbf{p}_{\max}\) are written as follows.

\begin{equation}
\begin{aligned}
    \mathbf{p}_{\min} &= \left( 
        \left\lfloor \frac{\theta_c - \arctan\!\left(\tfrac{\sqrt{2}d}{2r_c}\right)}{\psi} \right\rfloor,\,
        \left\lfloor \frac{\phi_c - \arctan\!\left(\tfrac{\sqrt{2}d}{2r_c}\right)}{\psi} \right\rfloor
    \right), \\[6pt]
    \mathbf{p}_{\max} &= \left( 
        \left\lceil \frac{\theta_c + \arctan\!\left(\tfrac{\sqrt{2}d}{2r_c}\right)}{\psi} \right\rceil,\,
        \left\lceil \frac{\phi_c + \arctan\!\left(\tfrac{\sqrt{2}d}{2r_c}\right)}{\psi} \right\rceil
    \right).
\end{aligned}
\end{equation}
where $\lfloor \cdot \rfloor$ and $\lceil \cdot \rceil$ denote the floor (rounding down) and ceiling (rounding up) operators, respectively.

Then, for each pixel $\texttt{p}$ within this projection area, we compare the depth value \(r_c\) of boundary voxel \(\mathbf{c}\) with the depth value of the LiDAR return of this pixel. If the boundary voxel is closer to the sensor, it indicates a potential occlusion of the LiDAR point and is saved in pixel $\texttt{p}$. We denote the set of boundary voxels that saved in pixel $\texttt{p}$ as $\mathbf{T}$.

\begin{figure}[t] 
    \centering
    \includegraphics[width=\linewidth]{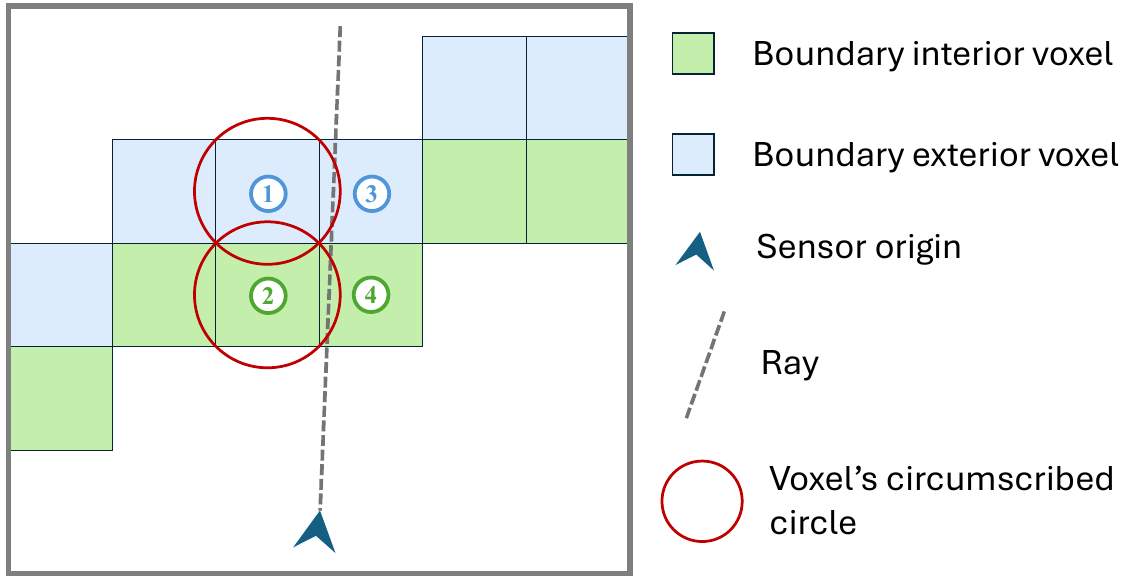}
    \vspace{-10pt} 

    \caption{The motivation for applying the slab method after the initial determination by depth-image rasterization. In this example, the ray intersects the circumscribed circles of boundary voxels no.1 and no.2, but does not actually pass through the voxels themselves.}

    \label{fig:depth_img}

    \vspace{-8pt}
\end{figure}

After the projection of all boundary voxels within the FoV, we retrieve the set of boundary voxels, denoted by $\mathbf{T}_i$ that {corresponding to} a pixel $\texttt{p}_i$ (Line 9). This set is intended to include the boundary voxels that may intersect the ray corresponding to the pixel $\texttt{p}_i$. However, due to the use of circumscribed-circle projection, the resulting candidate set is an over-approximation, and may contain voxels that are not actually intersected by the ray, as illustrated in Fig.~\ref{fig:depth_img}. To eliminate such false positives, we refine $\mathbf{T}_i$ using the slab method {(i.e., an algorithm used to solve the ray--box intersection problem)} \cite{Majercik2018Voxel} to perform an exact ray--voxel intersection determination. Voxels determined not to intersect the ray are subsequently removed from the set $\mathbf{T}_i$ (Line 11). After this process, \(\mathbf{T}_i\) comprises exactly the boundary voxels intersected by the corresponding LiDAR ray. Then, these boundary voxels in \(\mathbf{T}_i\) are sorted by their range to the sensor origin (i.e., their depth value) (Line 12).

Subsequently, we proceed to determine the ray segments that located outside the boundary surface layer, and traverse and update voxels on these segments to free state.
{Typically, the sensor origin lies in free space, i.e., within the interior of the boundary surface layer. Consequently, a ray emitted from the sensor origin initially lies within the interior of the boundary surface layer until it encounters boundary voxels, specifically those stored in \(\mathbf{T}_i\).}
We begin by checking the boundary voxels set \(\mathbf{T}_i\) from the beginning (with the lowest depth value). If a $\mathbf{b}_{\text{ext}}$ voxel (i.e., boundary exterior voxel) is reached, it indicates that the ray exits the boundary surface layer. Then, if we reach a $\mathbf{b}_{\text{int}}$ voxel (i.e., boundary interior voxel), it indicates that the ray re-enters the boundary surface layer and the ray segment between $\mathbf{b}_{\text{ext}}$ and $\mathbf{b}_{\text{int}}$ is at \textit{exterior} of the boundary surface layer. Thus, a ray casting traversal starts from the $\mathbf{b}_{\text{ext}}$ voxel and terminates at the $\mathbf{b}_{\text{int}}$ voxel. In the case where the ray casting traversal starts from the $\mathbf{b}_{\text{ext}}$ voxel and not encounters a $\mathbf{b}_{\text{int}}$ voxel, the ray casting traversal process is terminated when reaching the end point (i.e., reach the LiDAR return or exceed the sensor sensing range). The traversed voxels are updated to free state and are added to a container $\mathbf{L}$ for subsequent boundary map updates. The above procedures is outlined in Alg.~\ref{alg:boundary_update} (Lines 13--23). {Finally, the voxel that the LiDAR return lies in is updated to occupied state, and is also added into $\mathbf{L}$.}

Fig.~\ref{fig:boundary_map}(e) illustrates an example of the proposed ray traversal strategy. For ray~3, the set of intersected boundary voxels is denoted as $\mathbf{T}_3 = \{\mathbf{c}_1, \mathbf{c}_2, \mathbf{c}_3, \mathbf{c}_4, \mathbf{c}_5, \mathbf{c}_6, \mathbf{c}_7\}$, ordered by increasing distance from the sensor origin.
Then, we examine voxels in $\mathbf{T}_3$. Among these voxels, $\mathbf{c}_3$ is identified as a boundary exterior voxel, indicating that the ray exits the boundary surface layer between $\mathbf{c}_2$ and $\mathbf{c}_3$, with the corresponding intersection point marked by a cross. Ray traversal is initiated at this location. The next boundary interior voxel, $\mathbf{c}_5$, indicates that the ray re-enters the boundary surface layer, and traversal is terminated accordingly. Subsequently, $\mathbf{c}_7$ is another boundary exterior voxel, implying that the ray exits the layer again, and traversal continues until the ray reaches the sensor sensing range. Consequently, in this example, ray traversal is performed on the segments between $\mathbf{c}_3$ and $\mathbf{c}_5$, and between $\mathbf{c}_7$ and the end point.

Note that $\mathbf{T}_i$ may be an empty set. This indicates that the ray does not intersect the boundary surface layer between the sensor origin and the end point. In this situation, the ray lies entirely inside the boundary surface layer. {Thus, no ray casting is required in this case.}

\begin{figure*}[t] 
    \centering
    \includegraphics[width=\textwidth]{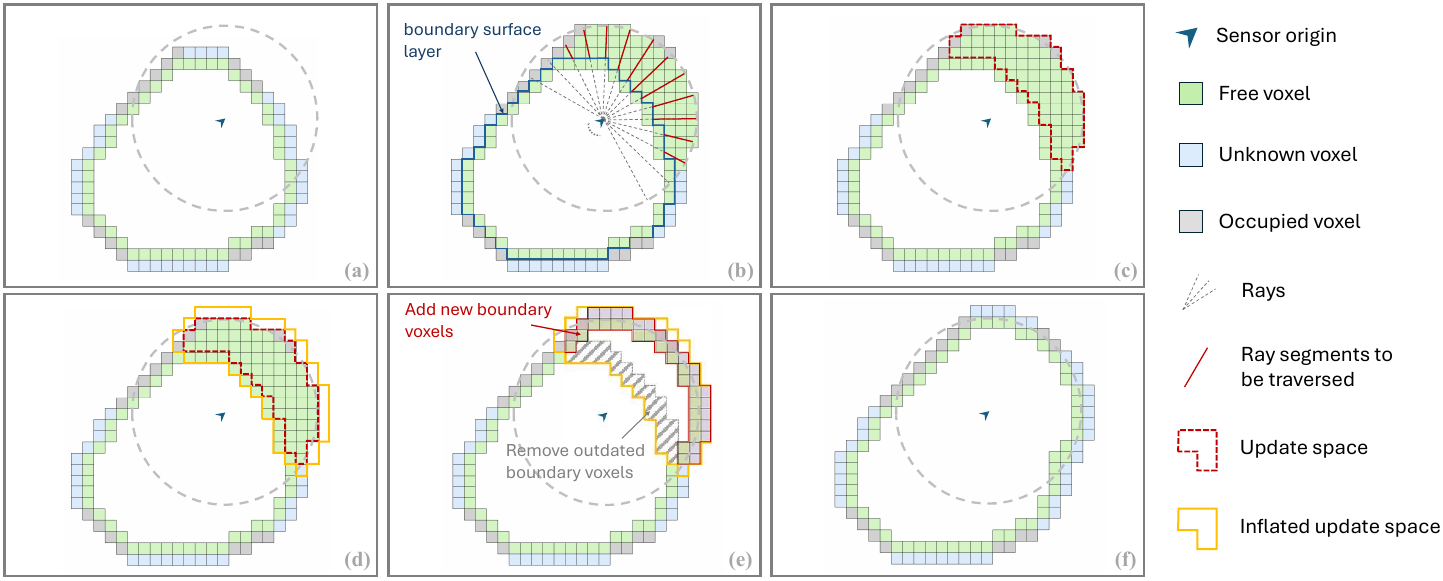}
    \vspace{-10pt} 

    \caption{(a) The boundary map before the update, where voxels with free, unknown and occupied state are colored as green, blue and grey, respectively. (b) Preform the truncated ray casting process, updating the voxels outside the boundary to free state. (c) Illustration of the update space, where voxels occupancy states change. (d) The inflated updated space, which is computed by the updated space by one voxel inflation. Voxels in the inflated updated space may have changes in their boundary voxel status. (e) Compute new boundary voxel status for all voxels in the inflated updated space. Remove outdated boundary voxels from and add new boundary voxels to the boundary map. (f) Illustration of the final updated boundary map.}

    \label{fig:upd_boundary_final}
\end{figure*}

\subsection{Direct Boundary Map Update}
\label{sec:upd_boundary}

In this section, we describe the complete procedure for updating the boundary map.
To begin, we present a brief introduction of the data structure used to maintain the boundary map, as this provides the foundation for understanding the update process. The boundary voxels are projected along the $z$-axis onto the $(x, y)$-plane. A hash-based 2D grid is then maintained on this plane, where each 2D grid cell stores the boundary voxels projected onto it. Within each cell, the boundary voxels are organized in an array. In other words, each grid cell in the hash-based 2D map maintains all boundary voxels that share the same $(x, y)$-coordinates as the cell. Note that the $z$-axis is chosen as the projection axis here for illustration; in practice, any of the three axes can serve this role. {This data structure is similar to BoundaryMap~\cite{BDM}, where a more detailed description is provided}.

We denote the sensor FoV as $\mathcal{V}$. The FoV projects onto the $(x,y)$-plane as a circular projection area, which we denoted as $\mathcal{V}^\text{2D}$ (Line 1).
To efficiently update the boundary map, we maintain a dynamic list of 2D grid cells, denoted as $\mathcal{K}$, corresponding to the cells of the 2D grid map within the area $\mathcal{V}^\text{2D}$.
When a new sensor measurement is received, its FoV $\mathcal{V}$ and corresponding circular projection area $\mathcal{V}^\text{2D}$ are also updated. This procedure is referred to as \texttt{UpdateCellsList} (Line 2). Then, 2D grid cells in $\mathcal{K}$ that no longer fall within the area $\mathcal{V}^\text{2D}$ are removed. Meanwhile, the 2D grid cells that “slide-in” the area $\mathcal{V}^\text{2D}$ are added into $\mathcal{K}$.
Next, for each 2D grid cell in $\mathcal{K}$, we retrieve the boundary voxels stored in that cell. These voxels are then checked against the current sensor FoV $\mathcal{V}$ to determine whether they lie within it. This completes the extraction of boundary voxels in the FoV from the boundary map (Line 3). 

After extracting the boundary voxels within the sensor FoV, we perform the truncated ray casting procedure (Section~\ref{sec:ray_cast}), which produces an update set $\mathbf{L}$ consisting of voxels whose occupancy states are modified by the new sensor measurements (Line~4). The update set $\mathbf{L}$ is then used to update the boundary map (Line~5), as detailed below. Based on $\mathbf{L}$, we define an inflated update space $\mathcal{F}$ that includes all voxels in $\mathbf{L}$ and their 6-connected neighbors (Lines 27--29). The notion of \emph{boundary status} refers to a label that specifies whether a voxel is a boundary voxel and, if so, its specific type. As discussed in~\cite{BDM}, only voxels inside $\mathcal{F}$ can have their boundary status updated, while those outside $\mathcal{F}$ remain unchanged.
The boundary map is updated incrementally by first processing the 2D grid cells in $\mathcal{K}$. For each cell, we access its stored boundary voxels, removing those that lie within the inflated update space $\mathcal{F}$ (Lines 30--34). Next, the boundary status of every voxel in $\mathcal{F}$ is re-computed. Newly identified boundary voxels are added into their respective 2D grid cells, similarly to~\cite{BDM}
, completing the boundary map update process (Lines 35--38).

\section{Experiments}
\label{sec:exps}
Extensive benchmark experiments were carried out to systematically assess the performance of the proposed mapping framework in comparison with several representative state-of-the-art approaches, including Uniform Grid~\cite{moravec1996robot}, OctoMap ~\cite{hornung2013octomap}, D-Map~\cite{cai2023occupancy} and BoundaryMap~\cite{BDM}. 
We conducted experiments on three public datasets.
{We use a diverse set of real-world, large-scale sequences to ensure a comprehensive evaluation across different sensing setups and motion patterns. Specifically, the first three sequences \textit{KAIST\_04}, \textit{Town\_03} and \textit{Roundabout\_01} are selected from HeLiPR dataset~\cite{helipr}. In addition, we include two widely used sequences from the KITTI dataset~\cite{geiger2013vision}, namely \textit{kitti\_00} and \textit{kitti\_02}. These sequences are all captured in real-world urban environments by vehicles equipped with high-resolution LiDAR sensors (e.g., 128-line and 64-line), providing large-scale and complex scenes for evaluation. To further enhance the diversity of the dataset in terms of platform and motion characteristics, we used a sequence from Newer College dataset~\cite{ramezani2020newer}, namely the \textit{college} sequence. Unlike the aforementioned vehicle-based data, this sequence is collected using a handheld device during a pedestrian traversal on a university campus, introducing distinct sensor motion patterns.}
More details of these sequences including the environment scale, travel distance and sensor type are listed in Table~\ref{tab:dataset}.

The experimental setup is described in Section~\ref{sec:setup}, followed by the
results, including the update time and memory consumption in Section~\ref{sec:exp_results}. {In addition, we validate the proposed method in a real-world task by using data collected from a long-range navigation experiment conducted in a multi-floor building (see Section~\ref{sec:real-world}).}

\begin{table}[t]
	\setlength{\tabcolsep}{3.0pt}
	\centering
	\caption{Details of datasets used in the benchmark experiments.}
	\label{tab:dataset}	
	\begin{threeparttable}
		\begin{tabular}{@{}ccccc@{}}
			\toprule
			Sequence & Environment Scale & Traveled & Sensor Type \\
            & (Bounding Box) ($\mathrm{m}^3$) & Distance ($\mathrm{m}$) & &  \\ \midrule
			\textit{KAIST\_04}      & $\mathrm{1,547\times1,275\times103}$ & 6,346   & 128-lines LiDAR \\
			\textit{Town\_03}     & $\mathrm{1,379\times1,741\times111}$ & 8,900   & 128-lines LiDAR \\
            \textit{Roundabout\_01}     & $\mathrm{1,702\times1,691\times135}$ & 9,035 & 128-lines LiDAR \\
			\textit{kitti\_00} & $\mathrm{653\times586\times58}$ & 3,724 & 64-lines LiDAR\\
            \textit{kitti\_02} & $\mathrm{1,035\times688\times94}$ & 5,067 & 64-lines LiDAR\\
            \textit{college} & $\mathrm{484\times483\times314}$ & 2,396 & 128-lines LiDAR\\
			\bottomrule
		\end{tabular}
	\end{threeparttable}
 \vspace{-0.2cm}
\end{table}

\subsection{Experiment Setup}
\label{sec:setup}
The experiments were conducted on a platform equipped with an Intel i7-1260P CPU, 64 GB of RAM.
The benchmark experiments were conducted across various map resolutions ranging from 0.5m to 0.1m.
In our experiments, the sensing range was set to $20\,\text{m}$. The size of the local map in BoundaryMap was set to $(40 \times 40 \times 40)\,\text{m}^3$ to fully contain the sensing range. Both our method and D-Map adopt similar depth-image rasterization technique, and the resolution of the depth image is configured by the LiDAR angular resolution, which is obtained by the user manual. 

\subsection{Results}
\label{sec:exp_results}

\begin{table*}[t]
	\centering
	\caption{Comparison of map update time (ms) at different resolutions.}
	\label{tab:update}
 \renewcommand*{\arraystretch}{1.0}
 \fontsize{8}{10}\selectfont 
	\begin{threeparttable}
		\begin{tabularx}{\textwidth}{@{}c 
		        >{\centering\arraybackslash}X
		        >{\centering\arraybackslash}X
		        >{\centering\arraybackslash}X
		        >{\centering\arraybackslash}X
		        >{\centering\arraybackslash}X
		        >{\centering\arraybackslash}X
		        >{\centering\arraybackslash}X
		        >{\centering\arraybackslash}X
		        >{\centering\arraybackslash}X
		        >{\centering\arraybackslash}X
		        >{\centering\arraybackslash}X
		        c@{}}
			\toprule
			\multicolumn{1}{l}{} & \multicolumn{4}{c}{\textit{KAIST\_04}} & \multicolumn{4}{c}{\textit{Town\_03}} & \multicolumn{4}{c}{\textit{Roundabout\_01}} \\
			\cmidrule(l){2-5} \cmidrule(l){6-9} \cmidrule(l){10-13}
			Resolution (m) & 0.5 & 0.3 & 0.2 & 0.1 & 0.5 & 0.3 & 0.2 & 0.1 & 0.5 & 0.3 & 0.2 & 0.1 \\
			\midrule
    Uniform Grid & 40.76 & 63.21 & $\times$ & $\times$ & 35.38 & 55.03 & $\times$ & $\times$ & 42.20 & $\times$ & $\times$ & $\times$ \\
    OctoMap & 318.05 & 521.19 & 766.74 & 1561.84 & 289.03 & 492.12 & 687.59 & 1399.87 & 325.85 & 547.45 & 813.13 & 1754.65 \\
    D-Map & 11.15 & 17.59 & 37.47 & 143.51 & 11.75 & 19.40 & 47.48 & 211.82 & \textbf{10.84} & 19.38 & 31.80 & 138.65 \\
    BoundaryMap & 42.28 & 71.22 & 113.70 & 406.04 & 36.56 & 60.40 & 92.11 & 306.60 & 44.20 & 74.93 & 122.32 & 382.98 \\
    Ours & \textbf{10.80} & \textbf{17.35} & \textbf{27.78} & \textbf{92.24} & \textbf{10.30} & \textbf{15.74} & \textbf{26.18} & \textbf{76.72} & {10.85} & \textbf{16.67} & \textbf{26.41} & \textbf{78.10} \\
    \midrule
		\end{tabularx}
	\end{threeparttable}	

	\begin{threeparttable}
		\begin{tabularx}{\textwidth}{@{}c 
		        >{\centering\arraybackslash}X
		        >{\centering\arraybackslash}X
		        >{\centering\arraybackslash}X
		        >{\centering\arraybackslash}X
		        >{\centering\arraybackslash}X
		        >{\centering\arraybackslash}X
		        >{\centering\arraybackslash}X
		        >{\centering\arraybackslash}X
		        >{\centering\arraybackslash}X
		        >{\centering\arraybackslash}X
		        >{\centering\arraybackslash}X
		        c@{}}
			\multicolumn{1}{l}{} & \multicolumn{4}{c}{\textit{kitti\_00}} & \multicolumn{4}{c}{\textit{kitti\_02}} & \multicolumn{4}{c}{\textit{college}} \\
			\cmidrule(l){2-5} \cmidrule(l){6-9} \cmidrule(l){10-13}
			Resolution (m) & 0.5 & 0.3 & 0.2 & 0.1 & 0.5 & 0.3 & 0.2 & 0.1 & 0.5 & 0.3 & 0.2 & 0.1 \\
			\midrule
    Uniform Grid & 28.34 & 43.39 & 63.12 & $\times$ & 28.07 & 42.44 & 64.19 & $\times$ & 20.83 & 30.46 & 45.50 & $\times$ \\
    OctoMap & 228.53 & 383.88 & 564.44 & 1102.95    & 241.45 & 382.50 & 529.96 & 1013.79 & 237.41 & 354.32 & 526.79 & 984.26 \\
    D-Map & 16.33 & 30.45 & 58.94 & 121.03          & 15.44 & 30.87 & 50.61 & 91.72 & 13.19 & 21.83 & 41.26 & 185.06 \\
    BoundaryMap & 29.62 & 46.96 & 71.14 & 201.01    & 28.53 & 46.06 & 70.33 & 175.42 & 22.07 & 35.83 & 59.25 & 171.0 \\
    Ours & \textbf{10.78} & \textbf{15.80} & \textbf{26.44} & \textbf{75.90} & \textbf{10.92} & \textbf{16.50} & \textbf{28.09} & \textbf{73.72} & \textbf{12.95} & \textbf{20.01} & \textbf{29.66} & \textbf{91.60}\\
    \bottomrule

        \end{tabularx}
	\end{threeparttable}
 \vspace{-0.3cm}    
\end{table*}

\begin{table*}[t]
	\centering
	\caption{Comparison of memory consumption (MB) at different resolutions.}
	\label{tab:memory}
 \renewcommand*{\arraystretch}{1.0}
 \fontsize{8}{10}\selectfont 
	\begin{threeparttable}
		\begin{tabularx}{\textwidth}{@{}c 
		        >{\centering\arraybackslash}X
		        >{\centering\arraybackslash}X
		        >{\centering\arraybackslash}X
		        >{\centering\arraybackslash}X
		        >{\centering\arraybackslash}X
		        >{\centering\arraybackslash}X
		        >{\centering\arraybackslash}X
		        >{\centering\arraybackslash}X
		        >{\centering\arraybackslash}X
		        >{\centering\arraybackslash}X
		        >{\centering\arraybackslash}X
		        c@{}}
			\toprule
			\multicolumn{1}{l}{} & \multicolumn{4}{c}{\textit{KAIST\_04}} & \multicolumn{4}{c}{\textit{Town\_03}} & \multicolumn{4}{c}{\textit{Roundabout\_01}} \\
			\cmidrule(l){2-5} \cmidrule(l){6-9} \cmidrule(l){10-13}
			Resolution (m) & 0.5 & 0.3 & 0.2 & 0.1 & 0.5 & 0.3 & 0.2 & 0.1 & 0.5 & 0.3 & 0.2 & 0.1 \\
			\midrule
    Uniform Grid & 6199.9 & 28703.4 & $\times$ & $\times$ & 8132.7 & 37651.5 & $\times$ & $\times$ & 11857.3 & $\times$ & $\times$ & $\times$ \\
    OctoMap & 96.5 & 549.3 & 1741.5 & 12588.1 & 147.2 & 712.8 & 2180.1 & 15300.6 & 150.6 & 869.3 & 2717.9 & 19752.5 \\
    D-Map & 295.8 & 804.3 & 2353.2 & 10610.0 & 398.4 & 1329.0 & 3199.7 & 14831.7 & 323.5 & 979.5 & 2427.4 & 10428.9 \\
    BoundaryMap & 40.5 & 135.4 & 316.3 & 1523.0 & 64.1 & 173.0 & 407.0 & 1935.5 & 67.2 & 180.6 & 422.8 & 1997.1 \\
    Ours & \textbf{38.9} & \textbf{127.3} & \textbf{288.2} & \textbf{1290.3} & \textbf{62.5} & \textbf{165.0} & \textbf{379.3} & \textbf{1706.9} & \textbf{65.6} & \textbf{172.6} & \textbf{394.6} & \textbf{1764.2} \\
    \midrule
		\end{tabularx}
	\end{threeparttable}	

    \begin{threeparttable}
		\begin{tabularx}{\textwidth}{@{}c 
		        >{\centering\arraybackslash}X
		        >{\centering\arraybackslash}X
		        >{\centering\arraybackslash}X
		        >{\centering\arraybackslash}X
		        >{\centering\arraybackslash}X
		        >{\centering\arraybackslash}X
		        >{\centering\arraybackslash}X
		        >{\centering\arraybackslash}X
		        >{\centering\arraybackslash}X
		        >{\centering\arraybackslash}X
		        >{\centering\arraybackslash}X
		        c@{}}
			\multicolumn{1}{l}{} & \multicolumn{4}{c}{\textit{kitti\_00}} & \multicolumn{4}{c}{\textit{kitti\_02}} & \multicolumn{4}{c}{\textit{college}} \\
			\cmidrule(l){2-5} \cmidrule(l){6-9} \cmidrule(l){10-13}
			Resolution (m) & 0.5 & 0.3 & 0.2 & 0.1 & 0.5 & 0.3 & 0.2 & 0.1 & 0.5 & 0.3 & 0.2 & 0.1 \\
			\midrule
    Uniform Grid & 677.3 & 3135.7 & 10583.0 & $\times$ & 2042.7 & 9457.0 & 31917.3 & $\times$ & 2240.1 & 10370.9 & 35001.9 & $\times$ \\
    OctoMap & 62.6 & 225.6 & 649.1 & 4202.2         & 81.9 & 307.7 & 891.3 & 5804.7 & 29.9 & 132.9 & 408.2 & 2814.7 \\
    D-Map & 179.7 & 586.0 & 1371.0 & 5948.9 & 182.1 & 487.5 & 1279.6 & 5619.1 & 68.0 & 230.7 & 697.0 & 3608.1 \\
    BoundaryMap & 30.9 & 84.0 & 200.2 & 989.4 & 38.3 & 128.6 & 298.2 & 1417.9 & 9.1 & 33.8 & 93.2 & 589.3 \\
    Ours & \textbf{29.2} & \textbf{75.8} & \textbf{171.8} & \textbf{756.1} & \textbf{36.5} & \textbf{119.9} & \textbf{268.6} & \textbf{1177.9} & \textbf{7.4} & \textbf{25.7} & \textbf{65.4} & \textbf{362.0} \\
    \bottomrule
		\end{tabularx}
	\end{threeparttable}
    
     \begin{tablenotes}
      \small
      \item \textit{Note:} $\times$ indicates that the method failed due to memory consumption exceeding the limit.
    \end{tablenotes}
    
 \vspace{-0.3cm}    
\end{table*}

\begin{table}[t]
	\setlength{\tabcolsep}{5.5pt}
	\centering
	\caption{Comparison of the traversed voxels of our method and classical ray casting at different scans.}
	\label{tab:traversed}	
	\begin{threeparttable}
		\begin{tabularx}{\linewidth}{@{}c 
		        >{\centering\arraybackslash}X
		        >{\centering\arraybackslash}X
		        >{\centering\arraybackslash}X
		        >{\centering\arraybackslash}X
		        c@{}}
			\toprule
			Scan No. & 1 & 2 & 3 & 5 & 20 \\
            \midrule
            Ours $(N_p)$ & 9,668,163 & 203,297 & 168,653 & 135,871 & 129,369 \\
            Classical $(N_r)$ & 9,668,163 & 9,588,663 & 9,488,954 & 9,461,390 & 9,180,961 \\
            Ratio $({N_p}/{N_r})$ & 100\% & 2.12\% & 1.78\% & 1.43\% & 1.41\%
 \\
			\bottomrule
		\end{tabularx}
	\end{threeparttable}
 \vspace{-0.4cm}
\end{table}

Table~\ref{tab:update} reports the map update time of our method and the baseline approaches. Our method consistently outperform all baselines in all sequences under all resolutions. Specifically, in \textit{Town\_03} sequence under map resolution of 0.1m, our method achieves a speed-up of 2.8 over D-Map, 4.0 times over BoundaryMap and 18.2 times over OctoMap{, while Uniform Grid failed due to exceeding memory}. These results highlight the efficiency of the proposed truncated ray casting method during map updates. To further illustrate the efficiency of this process, Table~\ref{tab:traversed} further compares the number of voxels traversed by truncated ray casting and normal ray casting. Both methods traverse the same voxels in the first scan. In subsequent scans, however, the truncated ray casting traverses substantially fewer voxels (e.g., approximately $2\%$), leading to significantly lower update cost.

Table~\ref{tab:memory} reports the memory consumption of our method and the baselines. Both our method and BoundaryMap achieve substantial memory savings compared with Uniform Grid, OctoMap, and D-Map, as both methods represent the environment using two-dimensional (2D) boundary voxels instead of explicitly maintaining a full three-dimensional (3D) volume. Moreover, our method further reduces memory usage compared with BoundaryMap, with the advantage becoming more pronounced at higher resolutions. This improvement results from eliminating the local 3D occupancy grid map, whose memory usage scales cubically with the map resolution. For example, at the map resolution of 0.1m, our method reduces memory consumption by \(89.7\%\) and \(87.8\%\) compared to OctoMap and D-Map, respectively, on the \textit{KAIST\_04} sequence. On the \textit{college} sequence at the same resolution, our method achieves a \(38.6\%\) reduction compared to BoundaryMap. In summary, the proposed method achieves substantial reductions in both update time and memory usage.

\begin{table}[t]
	\setlength{\tabcolsep}{5.5pt}
	\centering
	\caption{Map update time (ms) on the real-world task}
	\label{tab:real}	
	\begin{threeparttable}
		\begin{tabularx}{\linewidth}{@{}c 
		        >{\centering\arraybackslash}X
		        >{\centering\arraybackslash}X
		        >{\centering\arraybackslash}X
		        c@{}}
			\toprule
			Uniform~Grid & OctoMap & D-Map & BoundaryMap & Ours  \\
            \midrule
            23.95 & 106.19 & 28.88 & {22.04} & \textbf{15.58} \\
			\bottomrule
		\end{tabularx}
	\end{threeparttable}
 \vspace{-0.2cm}
\end{table}

\subsection{Evaluation on Real-World Data}
\label{sec:real-world}

\begin{figure}[ht] 
    \centering
    \includegraphics[width=\linewidth]{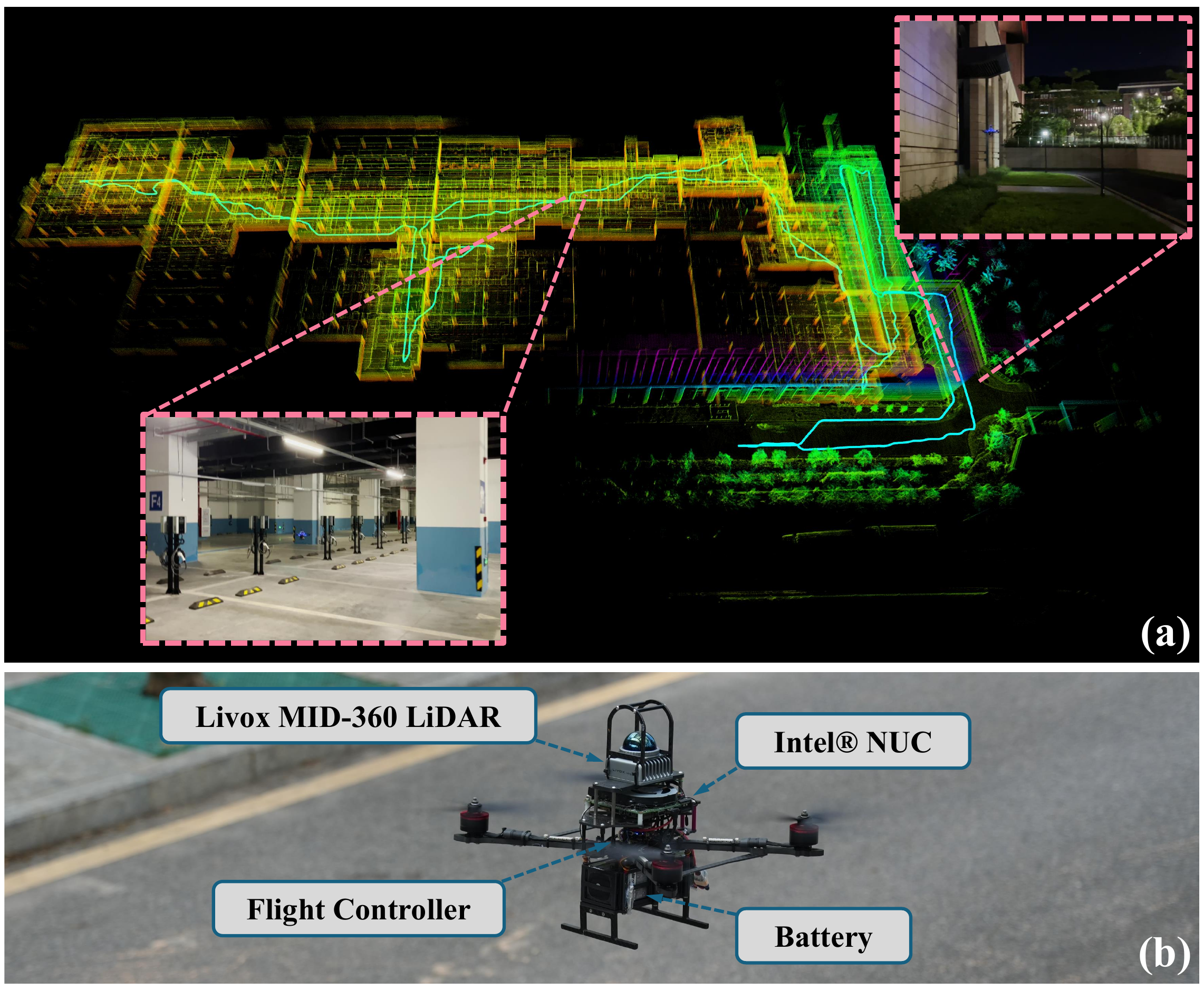}
    \vspace{-10pt} 

    \caption{(a) An illustration of the long-range navigation task in a multi-level building. (b) The Micro Aerial Vehicle (MAV) platform conducting the real-world flight.}

    \label{fig:real}
    \vspace{-10pt}
\end{figure}

{We compared the proposed method with BoundaryMap and several baseline approaches, including Uniform Grid, OctoMap, and D-Map, using real-world experimental data. In the experiment, a Micro Aerial Vehicle (MAV) performed long-range navigation in a multi-floor building. The resulting point cloud, the corresponding trajectory, and an illustration of the MAV platform are depicted in Fig.~\ref{fig:real}. The map resolution is set to 0.1m and sensor sensing range is 10m. Quantitative comparisons are summarized in Tab.~\ref{tab:real}. As shown, the proposed method achieves the best performance.}

\section{Conclusion}
\label{sec:conclude}
In this work, we presented an efficient occupancy mapping framework that builds upon the boundary representation. By exploiting geometric properties of the boundary, we introduce a truncated ray casting procedure that restricts voxel traversals to exterior regions only, thereby significantly reduces the map update time. Inherited from the low-dimensional boundary representation, our method exhibits significantly reduced memory consumption compared to traditional occupancy grid maps. In addition, the proposed direct boundary update scheme removes the dependency on a local 3D occupancy grid, which further reduce memory usage.

Extensive evaluations on multiple public datasets demonstrate that our approach delivers significant reductions in both update time and memory usage. Overall, the proposed method provides an efficient and scalable solution for real-time, large-scale occupancy mapping, making it well-suited for resource-constrained robotic platforms.

\bibliography{paper}

\end{document}